\documentclass[10pt,twocolumn,letterpaper]{article}

\usepackage{cvpr}
\usepackage{times}
\usepackage{epsfig}
\usepackage{graphicx}
\usepackage{algorithmic}
\usepackage{graphics}
\usepackage{epsfig}
\usepackage{amsmath,amssymb}
\usepackage{times}
\usepackage{epsfig, graphicx}
\usepackage{color}
\usepackage{xcolor}
\usepackage{amsmath,amssymb}
\usepackage{multirow}
\usepackage[ruled, linesnumbered, noend]{algorithm2e}
\usepackage{array}
\usepackage{slashbox}
\usepackage{subfigure}
\usepackage{units}
\graphicspath{{./figures/}}
\DeclareGraphicsExtensions{.pdf,.jpeg,.png,.jpg}

\usepackage[pagebackref=true,breaklinks=true,letterpaper=true,colorlinks,bookmarks=false]{hyperref}

\usepackage{xspace}

\usepackage{booktabs}

\def\eg{\emph{e.g}\onedot} 
\def\ie{\emph{i.e}\onedot}

\def\resp{\emph{resp}\onedot} 
\newcommand{\red}[1]{\textcolor{red}{#1}}

\usepackage[pagebackref=true,breaklinks=true,letterpaper=true,colorlinks,bookmarks=false]{hyperref}

\newcommand{\bfsection}[1]{\vspace*{0.1cm}\noindent\textbf{#1.}}
\cvprfinalcopy % *** Uncomment this line for the final submission

\def\cvprPaperID{5969} % *** Enter the CVPR Paper ID here

\ifcvprfinal\fi
\begin{document}

%%%%%%%%% TITLE
\title{Super-BPD: Super Boundary-to-Pixel Direction for Fast Image Segmentation}
%\title{{\Large Super-BPD: Super-Boundary-to-Pixel-Direction for Real-Time Image Segmentation}}

\author{
    Jianqiang Wan$^1$, Yang Liu$^1$,  Donglai Wei$^2$, Xiang Bai$^1$, Yongchao Xu$^1$\thanks{\scriptsize{Corresponding author}}~
    \\[2mm]
    $^1$Huazhong University of Science and Technology\ \ 
    $^2$Harvard University
    \\
   {\tt\small \{jianqw,yangl\_,xbai,yongchaoxu\}@hust.edu.cn,  donglai@seas.harvard.edu}
}

\maketitle
\thispagestyle{empty}

%%%%%%%%% ABSTRACT
\begin{abstract}
Image segmentation is a fundamental vision task and a crucial step for many applications. In this paper, we propose a fast image segmentation method based on a novel super boundary-to-pixel direction (super-BPD)
and a customized segmentation algorithm with super-BPD. Precisely, we define BPD on each pixel as a two-dimensional unit vector pointing from its nearest boundary to the pixel. In the BPD, nearby pixels from different regions have opposite directions departing from each other, and adjacent pixels in the same region have directions pointing to the other or each other (\ie, around medial points). We make use of such property to partition an image into super-BPDs, which are novel informative superpixels with robust direction similarity for fast grouping into segmentation regions. Extensive experimental results on BSDS500 and Pascal Context demonstrate the accuracy and efficiency of the proposed super-BPD in segmenting images. In practice, the proposed super-BPD achieves comparable or superior performance with MCG while running at $\sim$25fps vs. 0.07fps. Super-BPD also exhibits a noteworthy transferability to unseen scenes. The code is publicly available at \url{https://github.com/JianqiangWan/Super-BPD}.
\end{abstract}

%%%%%%%%% BODY TEXT
\section{Introduction}
\label{sec:introduction}
Image segmentation aims to decompose an image into non-overlapping regions, where pixels within each region share similar perceptual appearance, \eg, color, intensity, and texture. Image segmentation is a crucial step for many vision tasks such as object proposal generation~\cite{pont2017multiscale,zhang2017sequential}, object detection~\cite{juneja2013blocks}, semantic segmentation~\cite{farabet2013learning}. However, an efficient and accurate segmentation remains challenging. 

\begin{figure}
\centering
\includegraphics[width=\linewidth]{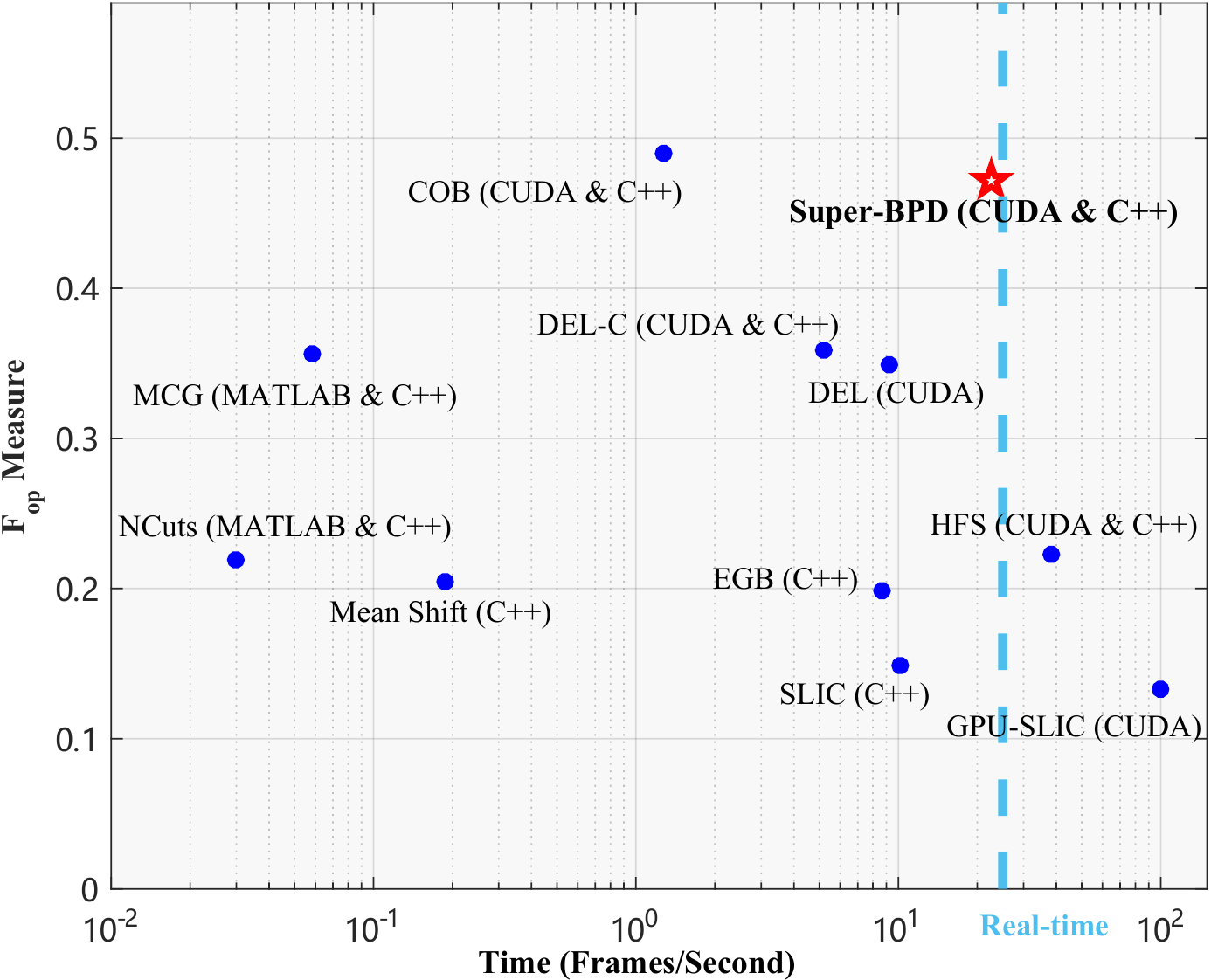}
\caption{Super-BPD achieves competing while near real-time performance on the PASCAL Context dataset~\cite{mottaghi2014role}. We plot the trade-off between efficiency and region F-measure accuracy ($F_{\text{op}}$)~\cite{pont2016supervised}.}
%\caption{Trade-off between efficiency and accuracy for some existing methods and the proposed super-BPD. Based on the region F-measure~\cite{pont2016supervised}, super-BPD achieves competing performance while being near real-time on the PASCAL Context Dataset~\cite{mottaghi2014role}.}
\label{fig:intro1}
\centering
\end{figure}

There are many unsupervised image segmentation methods that can be roughly categorized into early merging and clustering methods~\cite{zhu1996region,comaniciu2002mean}, active contours~\cite{kass1988snakes,caselles1997geodesic,chan2001active}, variational approaches~\cite{mumford1989optimal,sethian1999level}, watersheds~\cite{vincent1991watersheds,najman1996seg}, segmentation with graphical models~\cite{felzenszwalb2004efficient,shi2000normalized,boykov2001fast}. Though these classical methods are mathematically rigorous and achieve desirable results in some  applications, as depicted in Fig.~\ref{fig:intro1}, they usually do not perform well in segmenting natural images or are not very efficient. Superpixel segmentation~\cite{achanta2012slic,tu2018learning} is an efficient alternative that over-segments an image into small and compact regions. A grouping process~\cite{liu2018deep} is usually involved in producing final segmentation.   

Thanks to convolutional neural networks (CNNs), semantic segmentation~\cite{long2015fully,chen2018deeplab,zhao2017pyramid} that classifies each pixel into a predefined class category has witnessed significant progress in both accuracy and efficiency. Nevertheless, it does not generalize well to
unseen object categories. Alternatively, for image segmentation, some methods~\cite{arbelaez2011contour,pont2017multiscale,maninis2018convolutional,fang2019piecewise} resort to learn contours, followed by a transformation to bridge up the gap between contours and segmentation. As shown in Fig.~\ref{fig:intro1}, though these methods achieve impressive performances, the inevitable contour to segmentation transformation takes great effort to remedy leakage problems at weak boundaries and is usually time-consuming. 

%leverage classical machine learning or CNN to improve the performance. In particular, most learning-based methods~\cite{arbelaez2011contour,maninis2018convolutional,fang2019piecewise} first learn to detect contours, followed by a transformation from contours to segmentation results. Oriented watersheds and globalization via spectral clustering are usually adopted to bridge up the gap between contours and segmentation. 

\begin{figure}
\centering
\includegraphics[width=1\linewidth]{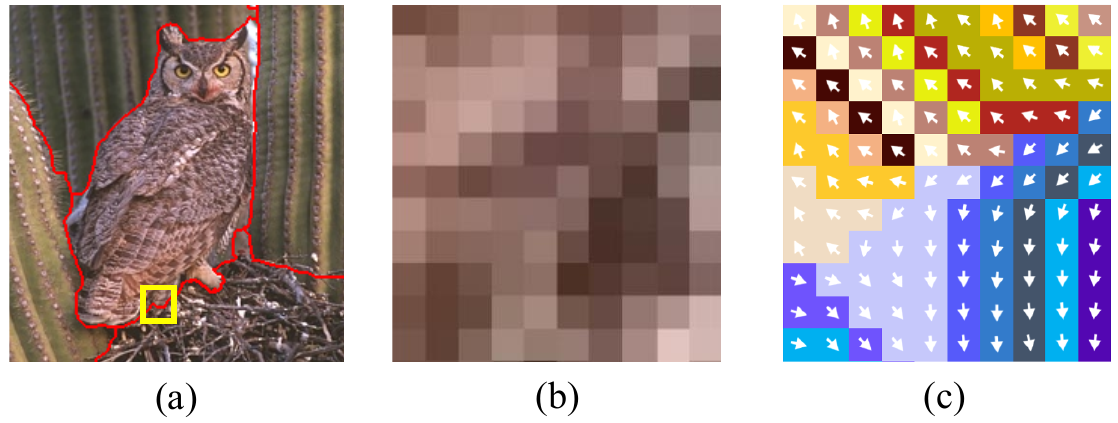}
\caption{Illustration of super-BPD results. (a) Given an image with super-BPD segmentation boundary (red), we zoom into a region with weak image boundaries (yellow). (b) Although pixels have similar values, (c) super-BPD can link pixels by the robustly predicted boundary-to-pixel direction (BPD), generating stripe-like segments on either side of the boundary for later grouping.}
%\caption{A segmentation example (with region boundaries in red) given by the proposed super-BPD. On the right side, we visualize the BPD (white arrow) on the yellow region with weak boundaries. Trajectories in different colors are different super-BPDs, having robust direction similarity for separating neighboring regions even with weak boundaries.}
\label{fig:example}
\centering
\end{figure}

Different from previous methods that directly learn contours and transform contours to segmentation, we propose a novel super boundary-to-pixel direction (super-BPD) and an efficient segmentation algorithm with super-BPD. Specifically, we introduce a boundary-to-pixel direction (BPD) on each pixel in terms of a two-dimensional unit vector, pointing from its nearest boundary to the underlying pixel.
The BPD not only provides contour positions but also encodes the relative position of each pixel to the corresponding region boundary and thus the relationship of neighboring pixels. This allows us to efficiently partition an image into super-BPDs such that each pixel and the pixel it points to and having similar directions are in the same super-BPD. The super-BPD can be regarded as a novel alternative of the classical superpixel, which provides robust direction for further grouping into segmentation regions. 

The set of super-BPDs form a region adjacency graph (RAG), where the edges are weighted by the direction similarity along the boundaries of adjacent super-BPDs. Nearby pixels within different regions have approximately opposite BPD, and hence small direction similarity. Such property also holds even at weak boundaries, where the learned BPD smoothly diverges to roughly opposite directions along the direction (see Fig.~\ref{fig:example} for an example). This equips the super-BPDs with robust direction similarity that helps to group similar super-BPDs within the same perceptual region and separate super-BPDs from different perceptual regions. We leverage such direction similarity between adjacent super-BPDs to partition the RAG into different clusters, resulting in a segmentation. As shown in Fig.~\ref{fig:intro1}, the proposed super-BPD achieves a good trade-off between accuracy and efficiency on PASCAL Context dataset~\cite{mottaghi2014role}. 
%BPDF also demonstrates an appealing transferability to unseen images.

% An advantage of such BPDF is that it encodes the attractive/repulsive relationship between neighboring pixels, which allows for fast post-processing to produce a final segmentation output. Besides, as shown in Fig.~\ref{fig:example}, even on the weak boundaries, BPDF still provides accurate cues for separating nearby distinct regions. Indeed, we can walk along the direction for a few steps, providing robust direction similarity for superpixels grouping, leading to accurate segmentation. 

The main contribution of this paper is two-fold: 1) We present a novel super boundary-to-pixel direction (super-BPD), which is a powerful alternative of the classical superpixel. super-BPD provides robust direction similarity between adjacent super-BPDs, which allows for efficient image segmentation. 2) We propose an efficient segmentation algorithm with super-BPDs in a coarse-to-fine way based on the direction similarity, leading to a good trade-off between segmentation accuracy and efficiency.
%Different from segmentation from contours, such processing is very efficient; 
%3) The proposed DFSEG achieves a good trade-off between accuracy and efficiency in image segmentation on both PASCAL Context and BSDS500 datasets, and feature a noteworthy transferability to unseen images.

% The rest of this paper is organized as follows. We shortly review some related works in Sec.~\ref{sec:relatedwork}, followed by the proposed method in Sec.~\ref{sec:method}. We then conduct extensive experiments in Sec.~\ref{sec:experiments}. We then conclude and give some perspectives in~\refsec{sec:conclusion}.

\section{Related Work}
\label{sec:relatedwork}
We shortly review some works on image segmentation and other vision tasks leveraging direction information. 

\subsection{Image Segmentation}
\label{subsec:imaseg}
%We divide existing image segmentation methods into unsupervised and supervised methods.

\bfsection{Unsupervised Methods} Many image segmentation methods have been proposed in the past two decades and can be roughly classified into several categories. Early segmentation methods are driven by region merging and clustering methods. Typical examples are region competition~\cite{zhu1996region} and mean shift~\cite{comaniciu2002mean}. Active contours~\cite{kass1988snakes,caselles1997geodesic,chan2001active} are another type of popular segmentation methods that evolve region contours by minimizing some energy functions. Variational approaches~\cite{mumford1989optimal,sethian1999level} also attempt to minimize some energy functions based on some appropriate hypotheses about the underlying image (\eg, piece-wise constant in~\cite{mumford1989optimal}). A set of watersheds~\cite{vincent1991watersheds,najman1996seg} has been proposed from the community of mathematical morphology. They segment image domain into catchment basins (\ie, regions) and watershed lines (\ie, contours). Another popular family of segmentation methods is based on graphical models~\cite{shi2000normalized,boykov2001fast,felzenszwalb2004efficient}, which model image domain as a graph and attempt to cut graphs based on some energy minimization. Besides these segmentation methods, superpixel methods~\cite{achanta2012slic,ren2015gslicr} aim to over-segment an image into small and compact regions. 

%\medskip
\bfsection{Supervised Methods} %Though many unsupervised segmentation methods have been proposed and achieve desirable performances in some specific applications, they usually have difficulty in segmenting complex natural images. 
Many learning-based image segmentation methods have been proposed. Different from semantic segmentation that can be regarded as a pixel-wise category classification problem, the mainstream learning-based segmentation methods~\cite{arbelaez2011contour,pont2017multiscale,maninis2018convolutional,fang2019piecewise} start with learning contours. They then resort to oriented watershed transformations and globalization via spectral clustering to alleviate the leakage problem at weak boundaries. However, such a contour-to-segmentation process is usually time-consuming. In~\cite{wolf2018mutex}, the authors propose mutex watershed (MWS) by learning local attractive/repulsive affinities, followed by a modified maximum spanning tree to segment images. Another direction is to learn a feature embedding~\cite{liu2018deep} for SLIC superpixels~\cite{achanta2012slic}, where superpixels within the same region (\resp, different regions) have similar (\resp, very different) embedded features. A simple merging algorithm based on the embedded features is then adopted to group superpixels into perceptual regions.

The proposed super-BPD falls into supervised methods. Different from the existing learning-based methods, super-BPD does not rely on contours and is free of the time-consuming process to handle weak boundaries in transforming contours to segmentation. Super-BPD is a powerful alternative to classical superpixels. It provides robust direction similarity for efficiently grouping pixels within the same region, and separating nearby regions even with weak boundaries between them. This results in a good trade-off between accuracy and efficiency. Compared with~\cite{liu2018deep}, super-BPD does not require a separate superpixel generation and embedding step, more efficient to separate different nearby regions with weak boundaries.

\subsection{Direction Cues for Vision Applications}
The direction information has been recently explored in different vision tasks. Some methods rely on similar direction fields defined on regions of interest. For instance, deep watershed transform~\cite{bai2017deep} propose to learn the direction field on semantic segmentation and then regress the distance to boundaries based on the direction information, followed by a classical watershed to produce instance segmentation. Textfield~\cite{xu2019textfield} and DeepFlux~\cite{wang2019deepflux} defines a similar direction field on text areas and skeleton context for scene text detection and skeleton extraction, respectively. The direction cue is also explored in MaskLab~\cite{chen2018masklab} and IRnet~\cite{ahn2019weakly} for improving instance segmentation and weakly instance segmentation, respectively. PifPaf~\cite{kreiss2019pifpaf} and {PVNet}~\cite{peng2019pvnet} leverage direction cue for 2D human pose estimation and 6 DoF pose estimation, respectively. 

The proposed super-BPD builds upon boundary-to-pixel direction (BPD), which is similar to~\cite{bai2017deep,xu2019textfield,wang2019deepflux} but defined on the whole image domain instead of regions of interest. The BPD learning confirms that it is possible to learn the direction field encoding the relative position of each pixel with respect to region contours in natural images. In this sense, the BPD can be seen as an extension of the flux for binary object skeletonization~\cite{siddiqi2002hamilton} to natural images for image segmentation. super-BPD differs a lot with~\cite{bai2017deep, xu2019textfield,wang2019deepflux} in how to use BPD defined on the whole image. The major contribution is the extension of BPD to super-BPDs that enhances the robustness of direction information of BPD and consequently the robust direction similarity between neighboring super-BPDs. Based on it, we propose an efficient coarse-to-fine RAG partition algorithm, leading to an accurate and efficient generic image segmentation.

%in the purpose of using direction field, and more importantly in post-processing with the direction field. In fact, in this paper, we propose a novel post-processing method customized for the BPDF by leveraging the direction to achieve image segmentation of natural images.  

\section{Super Boundary-to-Pixel Direction}
\label{sec:method}

%\subsection{Overview}\label{subsec:overview}
%Most existing methods have difficulty in maintaining a good trade-off between the accuracy and efficiency of generic image segmentation. 

Current segmentation methods achieve great performances with time-consuming post-processing, which limits their usages in practice. Whereas efficient segmentation methods provide degenerated results. We propose to remedy this issue by introducing a novel super boundary-to-pixel direction (super-BPD). The boundary-to-pixel direction (BPD) is defined 
on each pixel $p$ as the two-dimensional unit vector pointing from its nearest boundary pixel $B_p$ to $p$. Such BPD encodes the relative position between each pixel $p$ and the region (containing $p$) boundary. We adopt a CNN to learn such BPD, which is then used to partition the image into super-BPDs, a powerful alternative of classical superpixels. Super-BPDs provides robust direction similarity between adjacent super-BPDs, thus allowing fast image segmentation by partitioning the region (\ie, super-BPD) adjacency graph (RAG).

\begin{figure}[t]
\centering
\includegraphics[width=\linewidth]{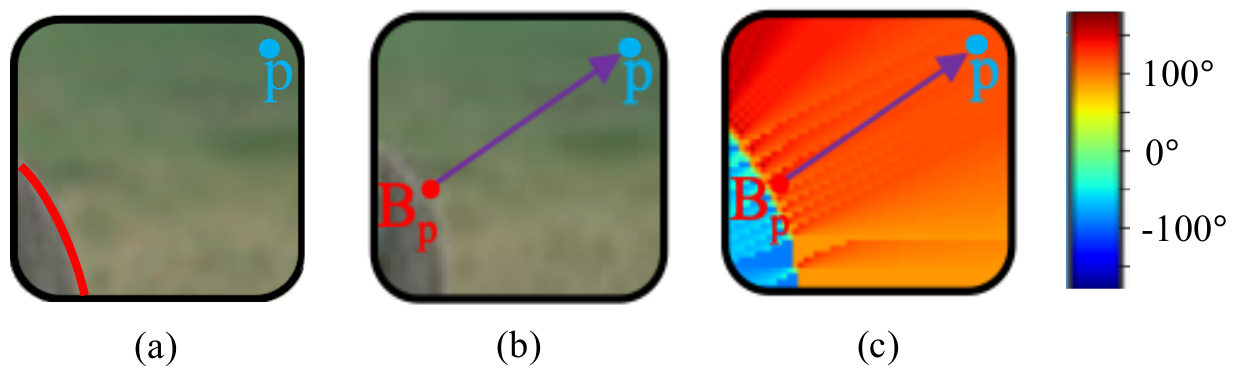}
\caption{Illustration of boundary-to-pixel direction (BPD). (a) For each pixel $p$ (ground truth boundary in red), (b) we find its nearest boundary pixel $B_p$. The BPD $\mathcal{D}_p$ is defined as the two-dimensional unit vector pointing from $B_p$ to $p$. (c) We predict BPD densely for each pixel and its direction is color-coded.}
\label{fig:bpd}
\centering
\end{figure}

%Specifically, we first adopt a CNN to learn the BPD in terms of a 2-channel map of direction field $\hat{D}$. We then bin the direction of $\hat{D}$ into 8 bins pointing to each of the 8 neighbors (C8 connectivity). Based on this, we partition the image into superpixels such that each pixel $p$ and the pixel $n_p$ that $p$ points to belong to the same superpixel if they share similar directions. We then construct a region (\ie, superpixel) adjacency graph (RAG), and weigh each edge linking two superpixels by the mean direction similarity of pixels around their boundaries. Based on the edge-weighted RAG, we develop a customized algorithm to group superpixels into regions. We argue and prove that such a pipeline of segmentation named Sup-BPD is effective and efficient (Fig.~\ref{fig:pipeline}).

%In particular, as shown in Fig.~\ref{fig:intro1}, the proposed method aptly named BPDF can separate two adjacent regions with weak boundaries. 

% \begin{figure}
% \centering
% \includegraphics[width=0.4\paperwidth]{./figures/representationofdirectionfieldcrop.pdf}
% \caption{Illustration of BPDF. For each pixel $p$, we find its nearest boundary pixel $B_p$ within different regions as $p$. The BPDF $D(p)$ is defined as the two-dimensional unit vector pointing from $B_p$ to $p$. On the right side, we visualize the direction of $D$.}
% \label{fig:defbpdf}
% \centering
% \end{figure}

\subsection{Boundary-to-Pixel Direction (BPD)}
\label{subsec:BPDF}
%Instead of explicitly detecting the boundary between adjacent regions by classifying each pixel into contour or non-contour, 

%We propose a novel representation that contains richer information than binary contour representation. Specifically, we transform a given image segmentation composed of non-overlapping regions into the super boundary-to-pixel direction on the whole image domain. 

\bfsection{Definition}
As shown in Fig.~\ref{fig:bpd}, for each pixel in the image domain $p \in \Omega$, we find its nearest boundary pixel $B_p$. Then, the BPD at pixel $p$, $\mathcal{D}_p$, is defined as a two-dimensional unit vector pointing from $B_p$ to $p$ given by
\begin{equation}
\mathcal{D}_p \, = \, \overrightarrow{B_p p}/|\overrightarrow{B_p p}|,
\label{eq:defbpdf}
\end{equation}
where $|\overrightarrow{B_p p}|$ is the distance between $B_p$ and $p$. The BPD provides cues about contour positions and relative position of each pixel $p$ to its region boundary. Note that generating BPD from ground-truth annotation could be efficiently achieved by the distance transform algorithm. 

%Nearby pixels from different regions and the same region have approximately opposite and similar directions, respectively. The BPDF not only encodes contour positions but also provides relative position of each pixel $p$ to its region boundary. 
%We leverage this property of the proposed BPDF to efficiently group pixels into regions. 

\begin{figure*}[ht]
\centering
\includegraphics[width=\linewidth]{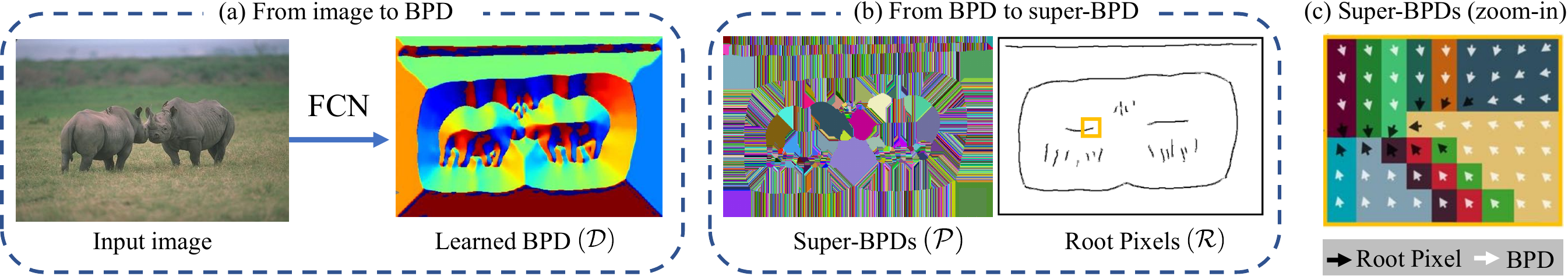}
\caption{Overview of super-BPD computation. (a) We adopt a Fully Convolutional Network (FCN) to learn the BPD field from the input image. (b) We then group BPDs into super-BPDs (colored regions) by the direction similarity threshold (Algo.~\ref{algo:bpdtosuper}) and extract root pixels. (c) We zoom in a region near the symmetry axis of the segment, where there are root pixels (black arrow) and regular BPDs (white arrow).}
%\includegraphics[width=0.76\paperwidth]{./figures/fromimagetobpdcrop.pdf}
%\caption{Illustration of the proposed BPD and super-BPD. For each pixel $p$, we find its nearest boundary pixel $B_p$ within different regions as $p$. The BPD $D(p)$ is defined as the two-dimensional unit vector pointing from $B_p$ to $p$. We adopt a Fully Convolutional Network (FCN) to learn the BPD, based on which we partition the image into super-BPDs represented by the root pixels. We then apply a dilation to merge nearby root pixels.}
\label{fig:defbpd}
\centering
\end{figure*}

% \begin{figure}
% \centering
% \includegraphics[width=0.4\paperwidth]{trendingofdepartcrop.pdf}
% \caption{Network architecture.}
% \label{fig:networkarchitecture}
% \centering
% \end{figure}

%\subsection{Network architecture}\label{subsec:network}
\bfsection{Architecture and Learning}
We adopt a Fully Convolutional Network (FCN) to predict BPD as a two-channel map with the same spatial size as the input image (Fig.~\ref{fig:defbpd}\red{a}). 
%The overall architecture of the adopted network is shown in~\ref{fig:networkarchitecture}. 
For a fair comparison with other methods, we adopt VGG16~\cite{vgg16network} as the backbone network, where the last max-pooling layer and all following layers are discarded. We also leverage ASPP layer~\cite{chen2018deeplab} to enlarge the receptive field, better coping with large regions. We extract features from different stages of VGG16 to aggregate multi-scale information. Specifically, we apply $1 \times 1$ convolutions to $conv3, conv4, conv5$, and ASPP layers, followed by a concatenation of these side output features after resizing them to the size of $conv3$. Finally, we apply three consecutive $1 \times 1$ convolutions on the fused feature maps, followed by an upsampling with bilinear interpolation to predict the BPD.

%\subsection{Loss Function}\label{subsec:objective}
We define the loss function in terms of both $L_2$-norm distance and angle distance for the BPD prediction $\hat{\mathcal{D}}$:
\begin{equation}
    L = \sum_{p \in \Omega} {w(p) ( \|\mathcal{D}_p- \hat{\mathcal{D}}_p\|^2 }+\alpha\|\cos ^{-1} \langle \mathcal{D}_p, \hat{\mathcal{D}}_p\rangle \|^{2} ),
    \label{eq:finalloss}
\end{equation}
where the adaptive weight at pixel $p$, $w(p) = 1/\sqrt{|GT_p|}$, is proportional to the inverse square root of the size of ground truth segment $GT_p$ containing $p$ and $\alpha$ is a hyper-parameter to balance the loss terms.
In practice, we set $\alpha=1$.

\begin{algorithm}[t]
\caption{Generate super-BPDs from the learned BPDs (Sec.~\ref{subsec:SBPDF}).}
\label{algo:bpdtosuper}
\SetKw{kAnd}{and}
\SetKw{kOr}{or}
\SetKw{kNot}{not}
\DontPrintSemicolon

\KwIn{Learned BPD ($\hat{\mathcal{D}}$), threshold $\theta_a$}
%\KwOut{Parent image $\mathcal{P}$ encoding super-BPDs}
\KwOut{Super-BPD ($\mathcal{P}$) and root pixel ($\mathcal{R}$)}
% \SetKwBlock{Begin}{function}{end function}
% \Begin($\text{FIND\_ROOT} {(}p{)}$)
% {
%  \eIf{$\mathcal{P}(p) = p$}{\Return{p}}{$\mathcal{P}(p) \gets \text{FIND\_ROOT}(p)$, \Return{$\text{FIND\_ROOT}{(}p{)}$}}
% }
\SetKwBlock{Begin}{function}{end function}
\Begin($\text{Get\_Super-BPDs} {(} \hat{\mathcal{D}}, \theta_a{)}$)
{
    // initialization \\% $\mathcal{P}$: parenthood, $R$: set of roots of superpixels \\
    $\mathcal{P} \gets p_0$, $\mathcal{R} \gets \emptyset$ \\
    // From BPD to super-BPD \\
    \ForEach{$p \in \Omega$}
    {
        \eIf{$\cos^{-1}\langle\hat{\mathcal{D}}_p, \hat{\mathcal{D}}_{n_p}\rangle \, < \, \theta_a$}{$\mathcal{P}(p) \gets n_p$}{$\mathcal{P}(p) \gets p$, $\mathcal{R} \gets p$}
    }
    
    \Return{$\mathcal{P}$, $\mathcal{R}$} \;
}
%\vspace{-4mm}
\end{algorithm}  

\subsection{BPD Grouping into Super-BPDs}
\label{subsec:SBPDF}
%We first partition the image into super-BPDs (\ie, superpixels with direction) and then merge nearby root pixels. 

%\bfsection{From BPD to super-BPD}
%From the learned BPD, we extract super-BPDs (encoded by a parent image $\mathcal{P}$), the stripe-like segments of different colors, and the root pixels of super-BPDs ($\mathcal{R}$) (Fig.~\ref{fig:defbpd}\red{b}). 
From the learned BPD, we extract super-BPDs, stripe-like segments encoded by a parent image $\mathcal{P}$, and their root pixels $\mathcal{R}$ close to regions' symmetry axes (Fig.~\ref{fig:defbpd}\red{b}). 

Precisely, inspired by the algorithms of computing component trees~\cite{salembier1998antiextensive,najman2006building,carlinet2014comparative}, we adopt a parent image $\mathcal{P}$ to encode the relationship between neighboring pixels.
Initially, the parent of each pixel $p$ is set to itself, $\mathcal{P}(p)=p$ and the set of root pixels $\mathcal{R}$ is empty.
For each pixel $p$, we define its next pixel $n_p$ as the neighboring pixel that is pointed to by $\hat{\mathcal{D}}_p$.
As depicted in Algo.~\ref{algo:bpdtosuper} (line 5-9), for each pixel in the raster order, if the angle between its BPD and that of its next pixel $n_p$, $\hat{\mathcal{D}}_p$ and $\hat{\mathcal{D}}_{n_p}$, is smaller than a given threshold $\theta_a$, we group them together by setting $\mathcal{P}(p)$, the parent of $p$, to $n_p$. Otherwise, we insert $p$ into the set of root pixels $\mathcal{R}$.
%With this, we partition the image into a forest of trees, indexed by root pixels in $\mathcal{R}$.
The final parent image $\mathcal{P}$ partitions the image into a forest of trees, each of which is rooted at a root pixel in $\mathcal{R}$.
We define each tree as a super-BPD.

Following the definition of BPD in Sec.~\ref{subsec:BPDF}, the direction around boundary pixels departs from each other, forming repulsive edges for separating neighboring pixels of different regions (even at weak boundaries, see Fig.~\ref{fig:example}).
% making the pixels around boundary present close to the leaf side of each tree. 
Pixels near the symmetry axis of each region also have approximately opposite directions, resulting in root pixels. Therefore, as shown in Fig.~\ref{fig:defbpd}\red{b}, 
the root pixels lie close to the symmetry axis of each region, allowing a safe merging of nearby root pixels that are usually within the same region.

\section{Image Segmentation with Super-BPD}
\label{subsec:BPDFtoseg}
\begin{figure}
\centering
\includegraphics[width=\linewidth]{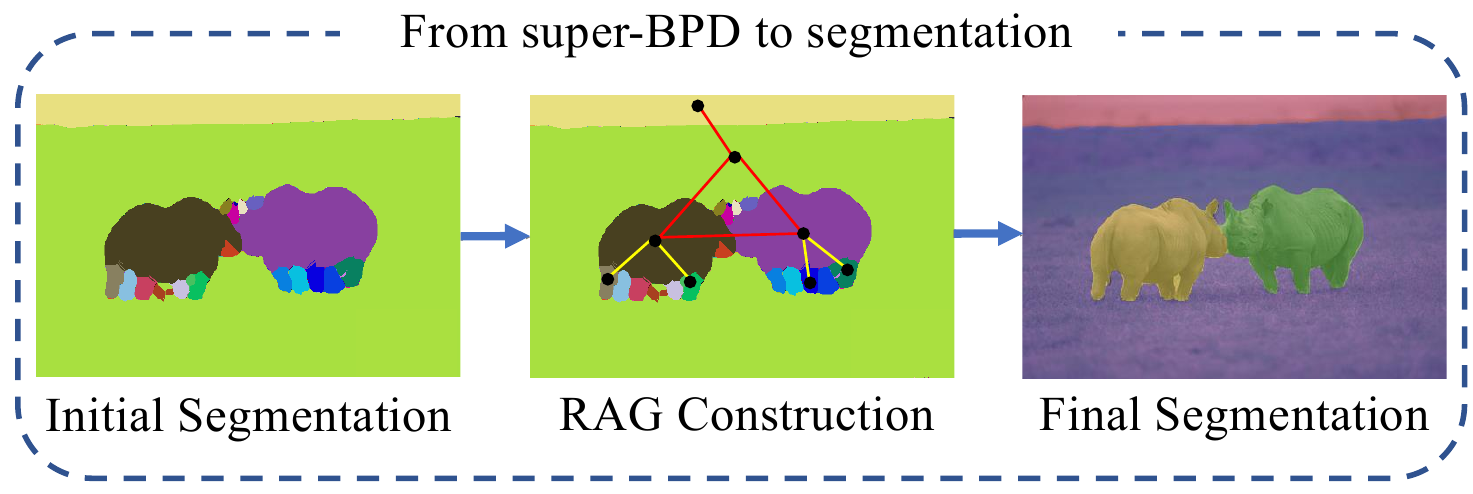}
\caption{Image segmentation from initial segmentation. We construct a RAG on the set of initial segments, and compute the size of each initial segment and the graph edge weight. We then partition the edge-weighted RAG into perceptual regions based on repulsive and attractive edges following the direction similarity.}
\label{fig:bpd2seg}
\centering
\end{figure}

%Similar to superpixel-based methods, we take a graph-based approach for image segmentation with super-BPD. 
We first obtain an initial segmentation via merging super-BPDs with nearby root pixels. Then, we construct an adjacency graph of regions (RAG) and apply the graph partitioning to merge initial segments as the final result (Fig.~\ref{fig:bpd2seg}).

% We propose a customized post-processing algorithm to transform the learned BPDF $\hat{D}$ into segmentation. Specifically,
% we first partition the image into superpixels following the direction information of $\hat{D}$. A region adjacency graph (RAG) is then constructed with edges weighted by the direction similarity $\mathcal{S}$ along the boundaries of neighboring superpixels. We then progressively group large, small, and tiny superpixels following $\mathcal{S}$, while ensuring that neighboring superpixels with small $\mathcal{S}$ (\ie, repulsive edges) are not grouped together. 
% The whole process is depicted in Fig.~\ref{fig:pipeline} and Algorithm~\ref{algo:bpdftoseg}, and detailed in the following.
%\medskip
%\medskip

\bfsection{Initial Segmentation}
As described above, root pixels of super-BPDs within the same ground truth region tend to be close to each other near the region's symmetry axis (Fig.~\ref{fig:defbpd}\red{b}).
%Thus, we can merge super-BPDs with neighboring root pixels to generate the initial segmentation.
%Since a pair of neighboring pixels on different sides of  have very different directions pointing to each other, there are many nearby roots representing adjacent super-BPDs within the same region (Fig.~\ref{fig:defbpd}\red{b}). 
Thus, we apply a simple dilation to group nearby root pixels and their corresponding super-BPDs to generate the initial segmentation. 
As depicted in Algo.~\ref{algo:bpdftoseg} (line 3-6), 
for each root pixel $r$, we update its parent $\mathcal{P}(r)$ to the last root pixel within the bottom half of $3 \times 3$ window $\mathcal{N}_3^b$ centered at $r$. We also update $\mathcal{R}$ by removing the merged root pixels.
With this simple method, we can group super-BPDs into a reasonable initial segmentation (Fig.~\ref{fig:bpd2seg}, left).

\begin{algorithm}[t]
\caption{Generate image segmentation from super-BPDs (Sec.~\ref{subsec:BPDFtoseg}).}
\label{algo:bpdftoseg}
\SetKw{kAnd}{and}
\SetKw{kOr}{or}
\SetKw{kNot}{not}
\DontPrintSemicolon

\KwIn{$\hat{\mathcal{D}}$, $\mathcal{P}$, $\mathcal{R}$, threshold $\theta_l$, $\theta_s$, $a_t$, $a_s$}
\KwOut{Set of linking edges $E_l$}
% \SetKwBlock{Begin}{function}{end function}
% \Begin($\text{FIND\_ROOT} {(}p{)}$)
% {
%  \eIf{$\mathcal{P}(p) = p$}{\Return{p}}{$\mathcal{P}(p) \gets \text{FIND\_ROOT}(p)$, \Return{$\text{FIND\_ROOT}{(}p{)}$}}
% }
\SetKwBlock{Begin}{function}{end function}
\Begin($\text{Super\_BPD2SEG} {(} \hat{\mathcal{D}}, \mathcal{P}, \mathcal{R}, \theta_l, \theta_s, a_t, a_s {)}$)
{
    // Initial segmentation \\
    \ForEach{$r \in \mathcal{R}$}
    {
        // merge nearby root pixels of super-BPDs \\
        \ForEach{$q \in \mathcal{N}_3^b(r)$}
        {
            \lIf{$q \in \mathcal{R}$}{$\mathcal{P}(r) \gets q$, $\mathcal{R}$.pop(r)}
        }
    }
    \;
    // Region adjacency graph construction\\
    %// initialization \\% $\mathcal{P}$: parenthood, $R$: set of roots of superpixels \\
    
    $(E_A^{\downarrow}, \mathcal{S}, \mathcal{A}) \gets \mathrm{Get\_RAG}(\hat{\mathcal{D}}, \mathcal{P}, \mathcal{R})$ \\
    \;
    // Graph Partitioning\\
    $E_l \gets \emptyset$\\
    \ForEach{$e = (r_1, r_2) \in E_A^{\downarrow}$}
    {
        // merge similar large and small initial seg. \\
        \If{$\min(\mathcal{A}_{r_1}, \mathcal{A}_{r_2}) > a_t$ \kAnd \kNot Rep($r_1, r_2$) \kAnd $\mathcal{S}(e) > h_{\theta_l,\theta_s,a_s}(\mathcal{A}_{r_1}, \mathcal{A}_{r_2})$  }
        {
            %\If{\textbf{not} Rep($r_1, r_2$)}
            %{
            Merge($r_1, r_2$), $E_l$.push(e) //updating
            %}
        }
    }
    
    \ForEach{$e = (r_1, r_2) \in E_A^{\downarrow}$}
    {
        // merge tiny initial seg. \\
        \If{$\min(\mathcal{A}_{r_1},\mathcal{A}_{r_2}) < a_t$ \kAnd \kNot Rep($r_1, r_2$)}
        {
            %\If{\textbf{not} Rep($r_1, r_2$)}
            %{
            Merge($r_1, r_2$), $E_l$.push(e) //updating
            %}
        }
    }
    \Return{$E_l$} \;
}
%\vspace{-4mm}
\end{algorithm}  

\bfsection{Region Adjacency Graph Construction}
%Based on previous super-BPD partition encoded by the parent image $\mathcal{P}$, we construct a region adjacency graph ($V_R, E$), where $E$ stands for the set of edges linking the roots of adjacent super-BPDs (Algo.~\ref{algo:bpdftoseg}, line 9). 
We construct a region adjacency graph ($\mathcal{R}, E$) from the initial segmentation, where $E$ stands for the set of edges linking the root pixels of adjacent segments (Algo.~\ref{algo:bpdftoseg}, line 9). 

%\bfsection{Graph Edge Weight}
For each edge $e=(r_1, r_2)\in E$, linking two regions $R_1$ and $R_2$, we compute its direction similarity $\mathcal{S}$. Let $B(e) = \{(p_i^1, p_i^2)\}$ be the set of neighboring pixels along the boundaries such that $p_i^1 \in R_1$ and $p_i^2 \in R_2$, we define $\mathcal{S}(e)$, the direction similarity on $e$, as following:
\begin{equation}
    \mathcal{S}(e) \, = \, \pi - \frac{\sum_{i=1}^{|B(e)|} \cos^{-1}\langle\hat{\mathcal{D}}_{\mathcal{P}_s(p_i^1)}, \hat{\mathcal{D}}_{\mathcal{P}_s(p_i^2)}\rangle}{|B(e)|},
    \label{eq:similarity}
\end{equation}
where $|B(e)|$ denotes the number of boundary points between $R_1$ and $R_2$, and $\mathcal{P}_s(p)$ stands for the $s$-th step starting from the pixel $p$. For example, $s = 0$ refers to the pixel itself.
With a direction similarity threshold $\mathcal{S}_0=\frac{\pi}{18}$, we divide all edges into attractive edges $E_A^{\downarrow}$, sorted in decreasing order of direction similarity, and repulsive edges $E_R$.
We define $Rep$ on a pair of adjacent segments to measure if they are repulsive. The $Rep$ is updated during the following merging process.

For each root pixel $r \in \mathcal{R}$, we compute the area $\mathcal{A}_r$ of its underlying initial segment.
Similar to~\cite{felzenszwalb2004efficient}, we divide initial segments into large, small and tiny by the hyper-parameter area thresholds $a_s$ and $a_t$.

\bfsection{Graph Partitioning}
We first greedily merge adjacent initial segments with either large or small sizes based on the direction similarity $\mathcal{S}$ (Algo.~\ref{algo:bpdftoseg}, line 13-16). 
Following~\cite{felzenszwalb2004efficient}, we make direction similarity thresholds adaptive to the size of the initial segments.
For each edge $e=(r_1,r_2)$, linking adjacent initial segments $R_1$ and $R_2$, we use a piece-wise constant threshold function $h_{\theta_l,\theta_s,a_s}(\mathcal{A}_{r_1}, \mathcal{A}_{r_2})$ with values as $\theta_l$ if both $\mathcal{A}_{r_1}$ and $\mathcal{A}_{r_2}$ are large segments ($\mathcal{A}>a_s$), and as $\theta_s (<\theta_l)$ otherwise. 

\begin{align}
\mathrm{h}_{\theta_l,\theta_s,a_s}(\mathcal{A}_{r_1},\mathcal{A}_{r_2})&= \begin{cases} \theta_l & \text{if } \min(\mathcal{A}_{r_1},\mathcal{A}_{r_2}) \geq a_s \\
\theta_s  & \text{otherwise}
\end{cases}\label{eqn:threshold}
\end{align}
% \min(\mathcal{A}_{r_1},\mathcal{A}_{r_2})<a_s
Then, we iterate through edges $e = (r_1, r_2)  \in E_A^\downarrow$ in the decreasing order of the direction similarity. 
We merge $R_1$ and $R_2$, if the direction similarity $\mathcal{S}(e)$ is larger than $h_{\theta_l,\theta_s,a_s}(\mathcal{A}_{r_1}, \mathcal{A}_{r_2})$ and the merge of  
$R_1$ and $R_2$ does not violate the repulsive rule, $Rep$.  
Note that each merge operation triggers the update of repulsive information $Rep$ between super-BPDs and super-BPD size $\mathcal{A}$. 

Finally, we merge initial segments with tiny sizes ($\mathcal{A}<a_t$) to their non-repulsive neighbors (Algo.~\ref{algo:bpdftoseg}, line 17-20). With this, we can clean up small crumb regions in the initial segmentation for better qualitative and quantitative results.

\bfsection{Runtime Analysis}
The whole segmentation from BPD is composed of three stages: 1) Super-BPD partition (Algo.~\ref{algo:bpdtosuper}, line 5-9). The complexity is $O(N)$, where $N$ denotes the number of pixels in image. 2) Nearby root pixels grouping (Algo.~\ref{algo:bpdftoseg}, line 3-6), which has a linear complexity with the number of root pixels. 3) Graph partition (Algo.~\ref{algo:bpdftoseg}, line 12-20), which has a quasi-linear time complexity with respect to the number of edges (\ie, in hundreds order) in RAG. Therefore, the whole post-processing has a near linear complexity, and is thus efficient. 

%Based on the candidate superpixels, we proposed a simple and effective method to achieve image segmentation. First of all, we get the relationships between each neighboring superpixel $\mathcal{N}_S(S_1,S_2)$, we define the average differences of each superpixel's boundary pixel $\mathcal{N}_p$ as difference $d_s$ between superpixels. Especially, we let the boundary pixels $(p_1,p_2)$ take $\sigma$ steps respectively along the way $( P(p_1), P(p_2) )$,and then we will get $\sigma$ different angles, we take the max of them as $(p_1, p_2)$'s difference $d_p$. Then we also discard some noisy superpixels whose areas are smaller than $\lambda_n$, for every neighboring superpixel $\mathcal{N}_s$ 

\section{Experiments}
\label{sec:experiments}
We conduct generic image segmentation experiments on PASCAL Context~\cite{mottaghi2014role} and BSDS500~\cite{arbelaez2011contour} dataset. %Following~\cite{liu2018deep}, we pre-train the network on Semantic Boundary Dataset (SBD)~\cite{hariharan2011semantic}, and then fine-tune on each benchmark dataset.

\begin{table*}[]
\small
\renewcommand\arraystretch{1.2}
\begin{center}
\begin{tabular}{c|l|c|c|c|c|c|c|c|c|c|c|c}
\toprule[1pt]
\multirow{2}{*}{Dataset} &\multirow{2}{*}{Methods} & \multicolumn{2}{c|}{$F_{b}$} & \multicolumn{2}{c|}{$F_{op}$} & \multicolumn{2}{c|}{Covering} & \multicolumn{2}{c|}{PRI} & \multicolumn{2}{c|}{VI} & \multirow{2}{*}{Time (s)} \\ \cline{3-12}
                                      && ODS         & OIS        & ODS        & OIS        & ODS        & OIS        & ODS         & OIS        & ODS         & OIS &\\\midrule[1pt]

&SLIC~\cite{achanta2012slic}           & 0.359      & 0.409       & 0.149        & 0.160      & -           & -      & -       & -          & -          & -          & 0.099    \\                                       
&Mean Shift~\cite{comaniciu2002mean}   & 0.397      & 0.406       & 0.204        & 0.214      & -           & -      & -       & -          & -          & -          & 5.320   \\ 
&MS-NCut~\cite{cour2005spectral}       & 0.380      & 0.429       & 0.219        & 0.285      & -           & -      & -       & -          & -          & -          & 33.40    \\
&EGB~\cite{felzenszwalb2004efficient}  & 0.432      & 0.454       & 0.198        & 0.203      & -           & -      & -       & -          & -          & -          & 0.116    \\ 
PASCAL&DEL~\cite{liu2018deep}                & 0.563      & 0.623       & 0.349        & 0.420      & 0.600       & 0.640  & 0.790   & 0.810      & 1.600      & 1.390      & 0.108    \\ 
Context&DEL-C~\cite{liu2018deep}              & 0.570      & 0.631       & 0.359        & 0.429      & 0.610       & 0.660  & 0.800   & 0.820      & 1.580      & 1.330      & 0.193    \\ 
&MCG~\cite{pont2017multiscale}         & 0.577      & 0.634       & 0.356        & 0.419      & 0.577       & 0.668  & 0.798   & 0.854      & 1.680      & 1.332      & 17.05    \\ 
&COB~\cite{maninis2018convolutional}   & \bf{0.755} & \bf{0.789}  & \bf{0.490}   & \bf{0.566} & \bf{0.739}  & \bf{0.803}  & \bf{0.878} & \bf{0.919} & \bf{1.150} & \bf{0.916}     & 0.790   \\ 
\cline{2-13}
&GPU-SLIC~\cite{ren2015gslicr}         & 0.322      & 0.340       & 0.133        & 0.157      & -           & -      & -       & -          & -          & -          & 0.010    \\ 
&HFS~\cite{cheng2016hfs}               & 0.472      & 0.495       & 0.223        & 0.231      & -           & -      & -       & -          & -          & -          & 0.026    \\ 
&Super-BPD \textbf{(Ours)}                          & \bf{0.704} & \bf{0.721}  & \bf{0.472}   & \bf{0.524} & \bf{0.730}  & \bf{0.770}  & \bf{0.880} & \bf{0.900} & \bf{1.150} & \bf{1.010}     & 0.044    \\
\bottomrule[1pt] 
& SLIC~\cite{achanta2012slic}           & 0.529       & 0.565      & 0.146      & 0.182      & 0.370      & 0.380      & 0.740       & 0.750      & 2.560       & 2.500      & 0.085 \\ 
&EGB~\cite{felzenszwalb2004efficient}  & 0.636       & 0.674      & 0.158      & 0.240      & 0.520      & 0.570      & 0.800       & 0.820      & 2.210       & 1.870      & 0.108 \\ 
&MS-NCut~\cite{cour2005spectral}       & 0.640       & 0.680      & 0.213      & 0.270      & 0.450      & 0.530      & 0.780       & 0.800      & 2.230       & 1.890      & 23.20 \\ 
&Mean Shift~\cite{comaniciu2002mean}   & 0.640       & 0.680      & 0.229      & 0.292      & 0.540      & 0.580      & 0.790       & 0.810      & 1.850       & 1.640      & 4.950 \\ 
&DEL~\cite{liu2018deep}                & 0.704       & 0.738      & 0.326      & 0.397      & 0.590      & 0.640      & 0.810       & 0.850      & 1.660       & 1.470      & 0.088\\ 
&DEL-C~\cite{liu2018deep}              & 0.715       & 0.745      & 0.333      & 0.402      & 0.600      & 0.660      & 0.830       & 0.860      & 1.640       & 1.440      & 0.165 \\
BSDS&MWS~\cite{wolf2018mutex}              & -           & -          & -          & -          & -          & -          & 0.826       & -          & 1.722       & -          & 0.580 \\ 
500&gPb-UCM~\cite{arbelaez2011contour} & 0.729       & 0.755      & 0.348      & 0.385      & 0.587      & 0.646      & 0.828       & 0.855      & 1.690       & 1.476      & 86.40 \\ 
&MCG~\cite{pont2017multiscale}         & \bf{0.744}  & \bf{0.777} & \bf{0.379} & \bf{0.433} & 0.613      & \bf{0.663} & 0.832       & \bf{0.861} & 1.568       & \bf{1.390} & 14.50 \\ 
&COB~\cite{maninis2018convolutional}   & \bf{0.782}  & \bf{0.808} & \bf{0.414} & \bf{0.464} & \bf{0.664} & \bf{0.712} & \bf{0.854}  & \bf{0.886} & \bf{1.380}  & \bf{1.222} & -     \\   
&GT*                                   & 0.732       & 0.732      & 0.463      & 0.463      & 0.690      & 0.690      & 0.860       & 0.860      & 1.180       & 1.180      & -\\ 
\cline{2-13}
&GPU-SLIC~\cite{ren2015gslicr}         & 0.522       & 0.547      & 0.085      & 0.132      & 0.340      & 0.370      & 0.730       & 0.750      & 2.950       & 2.810      & 0.007 \\ 
&HFS~\cite{cheng2016hfs}               & 0.652       & 0.686      & 0.249      & 0.272      & 0.560      & 0.610      & 0.810       & 0.840      & 1.870       & 1.680      & 0.024 \\ 
&Super-BPD \textbf{(Ours)}                          & 0.695       & 0.700      & 0.360      & 0.380      & \bf{0.640} & 0.650      & \bf{0.840}  & 0.850      & \bf{1.480}  & 1.430      & 0.036 \\
 \bottomrule[1pt] 
\end{tabular}
\end{center}
%\vspace{-1mm}
%\caption{Quantitative evaluation on the PASCAL Context and BSDS500 dataset.  Super-BPD achieves the state-of-the-art performance among real-time methods ($2^{\text{nd}}$ row) and remains competitive against non-real-time methods. GT* is the rough annotation following~\cite{liu2018deep}.}
\caption{In-dataset evaluation results. On both PASCAL Context and BSDS500 dataset, Super-BPD achieves the state-of-the-art performance among real-time methods ($2^{\text{nd}}$ row) and remains competitive against non-real-time methods. GT* is the rough annotation~\cite{liu2018deep}.}
\label{tab:quantitativeresults}
\end{table*}

\subsection{Implementation Details}
\label{subsec:implementation}
For training on BSDS500 dataset, we adopt the same data augmentation strategy used in~\cite{liu2018deep}. Specifically, the training images are rotated to 16 angles and flipped at each angle, then we crop the largest rectangle from the transformed images, yielding 9600 training images. Since PASCAL Context dataset has enough images, We only randomly flip images during training. The proposed network is initialized with the VGG16 model pretrained on ImageNet~\cite{deng2009imagenet} and optimized using ADAM~\cite{kinga2015adam}. Both models are trained for first 80k iterations with initial learning rate $10^{-5}$ for backbone layers and $10^{-4}$ for extra layers. We then decay the learning rates to $10^{-6}$ and $10^{-5}$, respectively, and continue to train the model for another 320k iterations on both BSDS500 PASCAL Context.

For the hyper-parameter settings, unless explicitly stated otherwise, we set $\theta_a$ to 45 for super-BPD partition. The threshold $a_s$ for small and $a_t$ for tiny regions are fixed to 1500 and 200. The step $s$ involved in computing the direction similarity in Eq.~\eqref{eq:similarity} is set to 3. The other two hyper-parameters $\theta_l$ and $\theta_s$ for merging large and small regions are tuned for the optimal dataset setting (ODS) on each dataset.  

%All models are pretrained on SBD dataset~\cite{hariharan2011semantic} for first 80k iterations with initial learning rate $10^{-5}$ for backbone layers and $10^{-4}$ for extra layers, then the learning rate is reduced to $10^{-6}$ and $10^{-5}$ respectively. We fine tune pretrained model on BSDS500 with 70k iterations and 160k for Pascal Context using the same learning rate strategy. 

The proposed super-BPD is implemented with PyTorch platform. All experiments are carried out on a workstation with an Intel Xeon 16-core CPU (3.5GHz), 64GB RAM, and a single Titan Xp GPU. Training on PASCAL Context using a batch size of 1 takes about 6 hours.

\begin{table*}[]
\small
\renewcommand\arraystretch{1.2}
\begin{center}
\begin{tabular}{cc|c|c|c|c|c|c|c}
\hline
        & \multicolumn{4}{c|}{PASCAL Context $\rightarrow$ BSDS500} & \multicolumn{4}{c}{BSDS500 $\rightarrow$ PASCAL Context} \\ \hline
\multicolumn{1}{c|}{Methods} & $F_{op}$ & Covering   & PRI      & VI      & $F_{op}$   & Covering    & PRI      & VI  \\ \hline
%\multicolumn{1}{c|}{DEL}      &  \textcolor{red}{?}    &   \textcolor{red}{?}     &  \textcolor{red}{?}     &  \textcolor{red}{?}    &   \textcolor{red}{?}    &  \textcolor{red}{?}      &  \textcolor{red}{?}     &  \textcolor{red}{?}    \\ \hline
\multicolumn{1}{c|}{DEL-C}    & 0.328    & 0.58       & 0.82     & 1.73    & 0.319      & 0.57        & 0.76     & 1.73    \\ \hline
\multicolumn{1}{c|}{Super-BPD}    & \bf{0.347}   & \bf{0.61}       & \bf{0.83}     & \bf{1.53}       & \bf{0.356}   & \bf{0.62}        & \bf{0.81}     & \bf{1.59}   \\ \hline
\end{tabular}
\end{center}
%\vspace{-1mm}
%\caption{Quantitative evaluation in terms of ODS metrics for super-BPD and DEL~\cite{liu2018deep} under cross-dataset validation.}
\caption{Cross-dataset evaluation results. We compare the performance of super-BPD and DEL~\cite{liu2018deep} with the ODS metrics.}
\label{tab:crossdatasetvalidation}
%\vspace{-4mm}
\end{table*}

\begin{figure}
\centering
\subfigure[Segmentation results on some images in PASCAL Context.]
{
%\vspace{-6mm}
\includegraphics[width=0.98\linewidth]{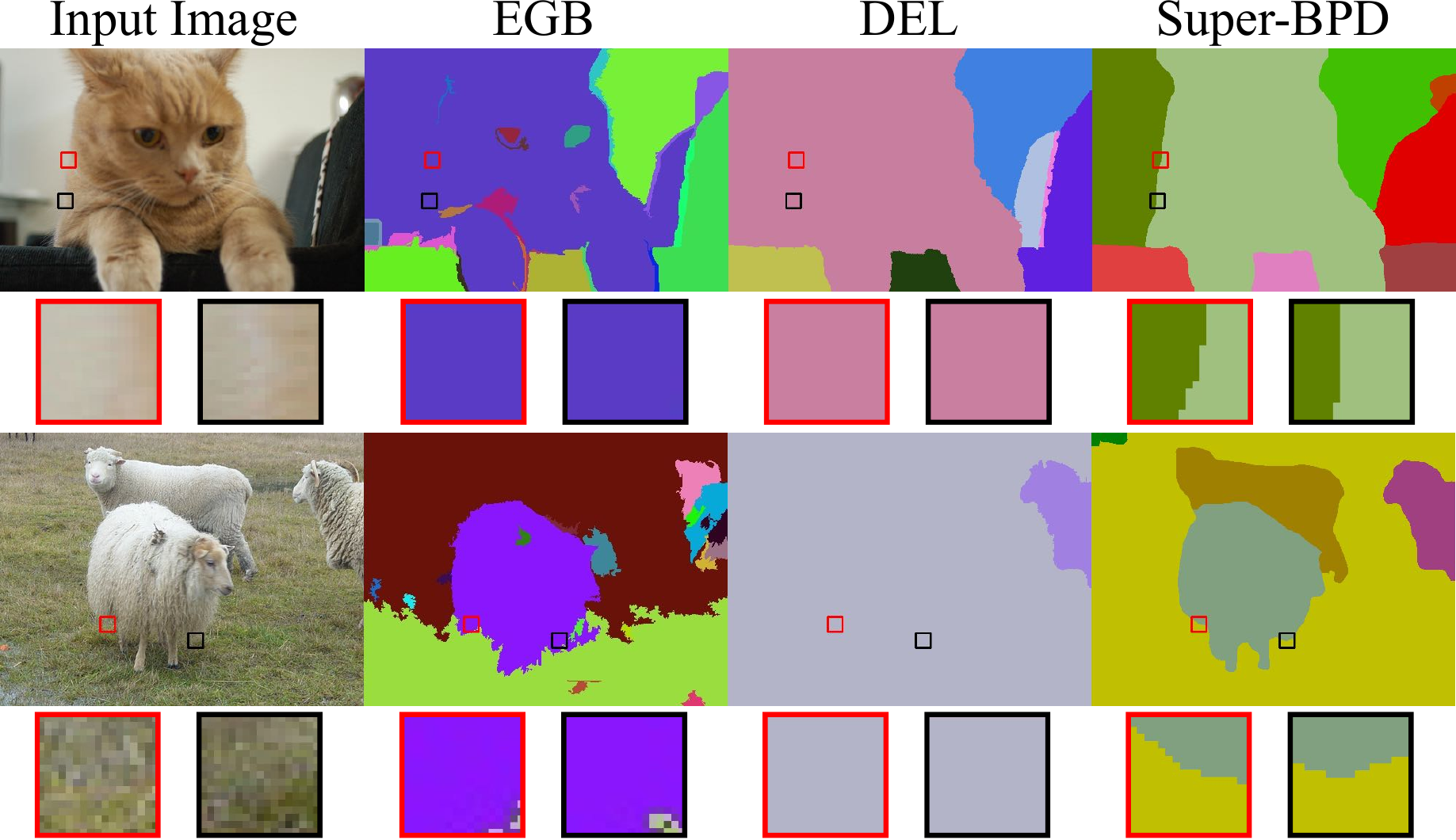}
    \label{fig:motivation_b}
}
\subfigure[Segmentation results on some images in BSDS500.]
{
\includegraphics[width=0.98\linewidth]{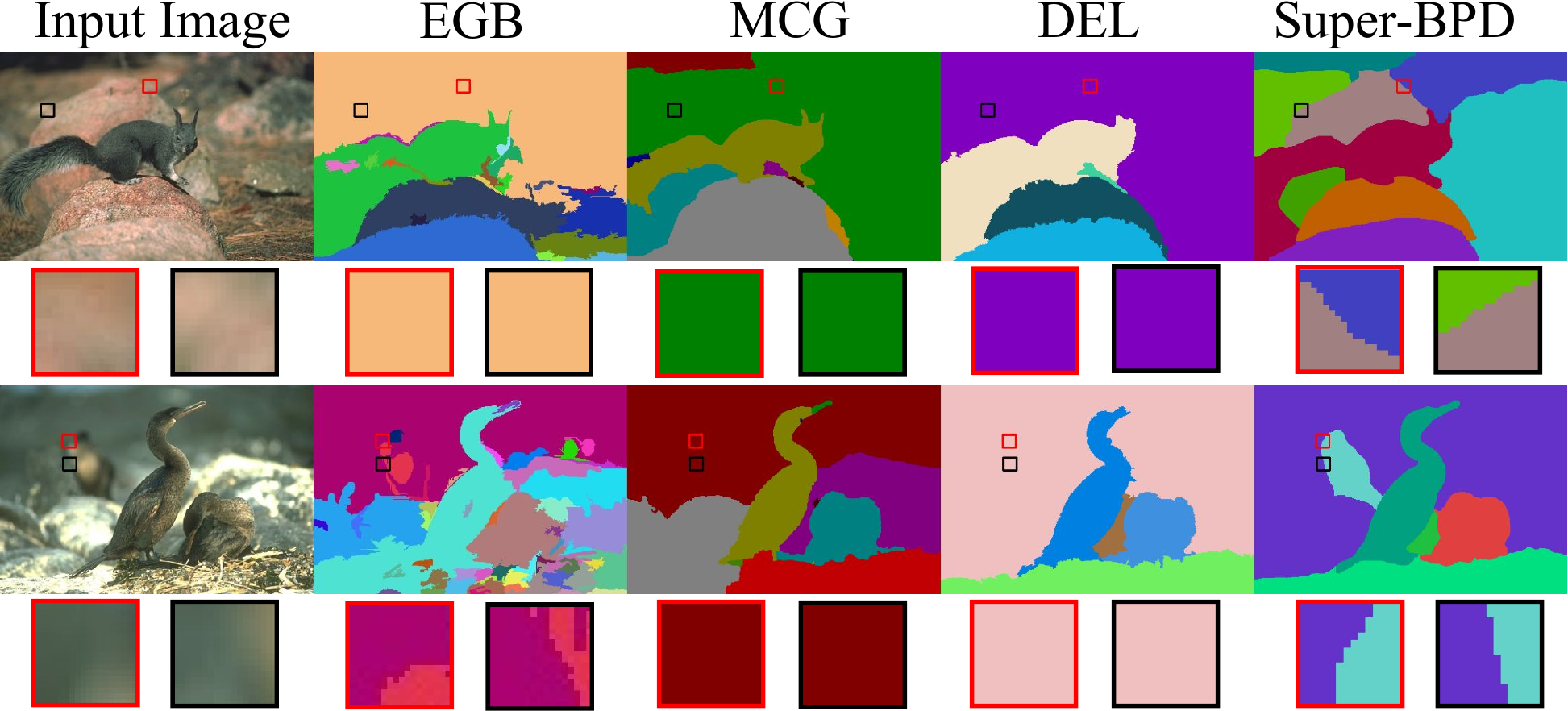}
\label{fig:bsdsqualitative}
}
\caption{Qualitative comparisons with other methods on some images from BSDS and PASCAL Context dataset. Super-BPD correctly segments natural images into perceptual regions despite existing weak boundaries.}
\label{fig:qualitativeresults}
\end{figure}

\subsection{Comparing with State-of-the-art Methods}
\label{subsec:results}
\bfsection{Datasets}
% \noindent\textbf{Semantic Boundary Dataset (SBD)}~\cite{hariharan2011semantic} contains 11355 images from the \textit{trainval} set of PASCAL VOC2011, it is divided into 8498 training and 2857 test images. This dataset contains annotations following the 20-class definitions in PASCAL VOC. We only use training images in the pre-train stage.
PASCAL Context~\cite{mottaghi2014role} contains precisely localized pixel-wise semantic annotations for the whole image. We ignore the semantics of each region for benchmarking the generic image segmentation methods. This dataset is composed of 7605 \textit{trainval} images and 2498 test images.

%object regions are used as ground-truth segmentation without semantic meaning for PASCAL Context dataset.
%\medskip
BSDS500~\cite{arbelaez2011contour} is a benchmark dataset for image segmentation and boundary detection. It is divided into 200 training images, 100 val images, and 200 test images. Each image has 5-10 different segmentation ground-truth annotations.
 
%\medskip 
\bfsection{Metrics}
We use four standard benchmark measures adopted in~\cite{arbelaez2011contour}: $F$-measure for boundaries $F_{b}$, segmentation covering (Covering), Probabilistic Rand Index (PRI), and Variation of Information (VI). We also evaluate the proposed Sup-BPD using another widely adopted metric $F$-measure for objects and parts $F_{op}$ introduced in~\cite{pont2016supervised}. 

\bfsection{In-Dataset Evaluation}
We compare the proposed super-BPD with other state-of-the-art image segmentation methods on both PASCAL Context and BSDS500. Some qualitative comparisons are shown in Fig.~\ref{fig:qualitativeresults}. Super-BPD correctly segments images into perceptual regions on both datasets.

%\bfsection{Quantitative Results}
The quantitative comparison on PASCAL Context is depicted in Tab.~\ref{tab:quantitativeresults} (top). Super-BPD achieves a pleasant trade-off between accuracy and efficiency on segmenting images in PASCAL Context. Specifically, super-BPD performs competitively with COB~\cite{maninis2018convolutional} while being much faster based on all evaluation metrics. Compared with MCG~\cite{pont2017multiscale}, super-BPD records an improvement of 11.6\% in ODS $F_{op}$ while being much more efficient. Super-BPD improves DEL and DEL-C~\cite{liu2018deep} by 12.3\% and 11.3\% in ODS $F_{op}$, respectively. Besides, super-BPD runs faster than DEL~\cite{liu2018deep}.  
%The comparision results on PASCAL Context dataset are shown in \reftab{tab:quantitativeresultsonpascalcontext}. since ODS of $F_{op}$ is a very important metric for segmentation, we mainly compare ODS of $F_{op}$ $vs.$ running time of different methods in this section. 

The quantitative comparison on BSDS500 dataset is shown in Tab.~\ref{tab:quantitativeresults} (bottom). Super-BPD also achieves a good trade-off between segmentation accuracy and efficiency. Specifically, super-BPD achieves competitive or superior performance with COB~\cite{maninis2018convolutional}, MCG~\cite{pont2017multiscale}, gPb-UCM~\cite{arbelaez2011contour}, and MWS~\cite{wolf2018mutex} while being more efficient. For a fair comparison with DEL~\cite{liu2018deep}, we also use the roughest annotation of each image for training. Super-BPD outperforms DEL and DEL-C~\cite{liu2018deep} by 3.4\% and 2.7\% in ODS $F_{op}$, respectively. We also report the result of using the roughest annotation as segmentation denoted as GT*. Super-BPD approaches the ``upper bound'' of ground-truth segmentation. It is noteworthy to mention that the lower accuracy on BSDS500 than on PASCAL Context is due to  
inconsistent annotations between different subjects. 

\bfsection{Cross-Dataset Evaluation}
\label{subsec:cross}
To demonstrate the generalization ability of the proposed super-BPD, we evaluate the model trained on one dataset and test the trained model on another dataset. We mainly compare super-BPD with DEL-C~\cite{liu2018deep}, which is also dedicated for a good trade-off between accuracy and efficiency. As depicted in Tab.~\ref{tab:crossdatasetvalidation}, super-BPD is robust in generalizing to unseen datasets. Specifically, super-BPD outperforms DEL-C~\cite{liu2018deep} for both PASCAL Context to BSDS500 and BSDS500 to PASCAL Context setting. In fact, super-BPD even outperforms DEL-C~\cite{liu2018deep} properly trained on the corresponding training set. 

%We mainly conduct two experiments to verify the generalization performance of BPDF, namely PASCAL Context to BSDS500 and the reverse cross-dataset validation. We show our results in \reftab{tab:crossdatasetvalidation} with comparisons to ~\cite{liu2018deep} in terms of ODS. It can be seen that BPDF achieves better performance while being faster.

\bfsection{Runtime Analysis}
Super-BPD has three stages: BPD inference, super-BPD partition, and segmentation with super-BPD. BPD inference on the GPU using VGG16 takes on average 22 ms for a $390 \times 470$ PASCAL Context image, and super-BPD partition and segmentation with super-BPD require  on average 22 ms for a PASCAL Context image on the CPU. As depicted in Tab.~\ref{tab:quantitativeresults}, super-BPD is much more efficient than the other competing methods while achieving comparable or superior performance.

\subsection{Ablation Studies}
\label{subsec:ablation}
\bfsection{ASPP Module} We first study the effect of ASPP module that increases receptive field for coping with large regions. As shown in Tab.~\ref{tab:influenceofaspp}, when the ASPP module is not used, the performance slightly decreases on both PASCAL Context and BSDS500 dataset. 

\bfsection{Loss Functions} We study the effect of different loss functions to train the network on PASCAL Context dataset. As depicted in Tab.~\ref{tab:influenceoflossfunction}, both $L_2$ loss function and the angular domain loss function achieve reasonable results, the combination of them by Eq.~\eqref{eq:finalloss} further improves the performance, and can also boost the convergence in training. 

\bfsection{Number of Steps $s$} We evaluate the influence of steps $s$ in computing direction similarity along boundaries between adjacent superpixels. As shown in Tab.~\ref{tab:influenceofwalkingsteps}, this step parameter $s$ has an impact on the segmentation performance. The best result is achieved with $s = 3$. In fact, adjacent pixels within different regions may have slightly different directions, but the overall direction trend is divergent. This explains the benefit of using $s = 3$.

%We get the best results when steps equal to 3, which demonstrates that we need walk some steps to confirm the trending of direction field because sometimes the angle difference between the two sides of boundary is small but the overall trending is separate. 

\begin{table}[t]
\small
\renewcommand\arraystretch{1.2}
\begin{center}
\begin{tabular}{c|c|cccc}
\hline
Dataset        & ASPP          & $F_{op}$      \\ \hline
PASCAL Context &               & 0.454           \\ \hline
PASCAL Context & \checkmark    & \bf{0.472}      \\ \hline
BSDS500        &               & 0.348          \\ \hline
BSDS500        & \checkmark    & \bf{0.360}      \\ \hline
\end{tabular}
\end{center}
%\vspace{-1mm}
\caption{Effects of ASPP module on the performance in ODS $F_{op}$.}
\label{tab:influenceofaspp}
%\vspace{-4mm}
\end{table}

\begin{table}[t]
\small
\renewcommand\arraystretch{1.2}
\begin{center}
\begin{tabular}{c|c|c|c}
\hline
Dataset                         & norm loss   & angular loss & $F_{op}$   \\ \hline
\multirow{3}{*}{PASCAL Context} & \checkmark  &              & 0.454       \\ \cline{2-4} 
                                &             & \checkmark   & 0.465       \\ \cline{2-4} 
                                & \checkmark  & \checkmark   & \bf{0.472}    \\ \hline
\end{tabular}
\end{center}
%\vspace{-1mm}
\caption{Effects of loss functions on the performance in ODS $F_{op}$.}
\label{tab:influenceoflossfunction}
%\vspace{-4mm}
\end{table}

\begin{table}[t]
\small
\renewcommand\arraystretch{1.2}
\begin{center}
\begin{tabular}{c|c|c|c|c|c}
\hline
Dataset        & s=0   & s=1    & s=3        & s=5    & s=7 \\ \hline
PASCAL Context & 0.460 & 0.462  & \bf{0.472} & 0.465  & 0.457    \\ 
\hline
BSDS500        & 0.353 & 0.354  & \bf{0.360} & 0.350  & 0.339    \\ 
\hline
\end{tabular}
\end{center}
%\vspace{-1mm}
\caption{Effects of steps $s$ on the performance in ODS $F_{op}$.}
\label{tab:influenceofwalkingsteps}
%\vspace{-4mm}
\end{table}

\section{Application: Object Proposal Generation}
\label{sec:application}
%\subsection{Application to object proposal generation}
%\label{subsec:application}

The object proposal generation task is a prerequisite step for a number of mid-level and high-level vision tasks such as object detection~\cite{girshick2015fast}. 

Following DEL~\cite{liu2018deep}, we use BING~\cite{cheng2014bing} as the baseline object proposal generation method.
We further improve BING's results with the multi-thresholding straddling expansion method (MTSE)~\cite{chen2015improving}, referenced as M-BING. 
To compare generic segmentation methods, we replace the EGB segmentation results~\cite{felzenszwalb2004efficient} used in MTSE~\cite{chen2015improving} with DEL~\cite{liu2018deep} and our proposed super-BPD respectively. 

\begin{figure}[t] \centering
\includegraphics[width=0.8\columnwidth]{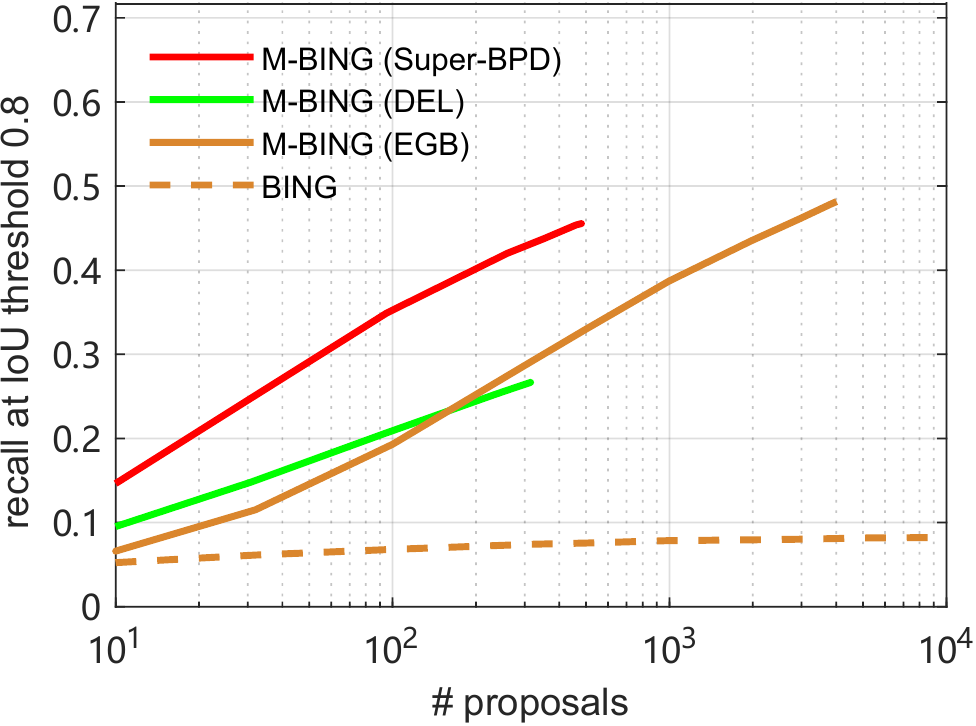}
\caption{Object proposal results on PASCAL VOC2007. Super-BPD outperforms other methods using the BING framework~\cite{cheng2014bing}.}
\label{fig:detectionrecallatiou08crop}
%\vspace{-4mm}
\end{figure}

On PASCAL VOC2007 test set, we plot the detection recall with 0.8 IoU overlap versus the number of proposals in Fig.~\ref{fig:detectionrecallatiou08crop}. 
M-BING with DEL has a slight improvement over M-BING until around 100 object proposals.
On the other hand, M-BING with super-BPD significantly improves upon M-BING by nearly twice until 500 proposals.

\begin{figure}[t] \centering
\includegraphics[width=0.98\columnwidth]{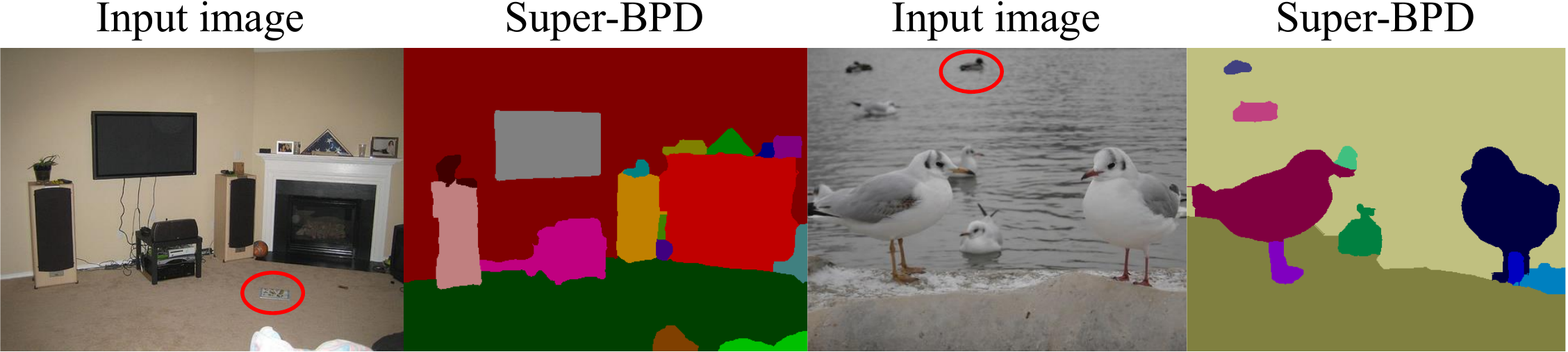}
\caption{Some failure cases. Super-BPD does not perform well in segmenting very small perceptual regions.}
\label{fig:failurecase}
%\vspace{-4mm}
\end{figure}

%predict correct direction around small regions due to the dramatic changes of direction.   

\section{Conclusion and Limitations}
\label{sec:conclusion}

We propose a fast image segmentation method based on a novel super boundary-to-pixel direction (super-BPD) and a customized segmentation with super-BPD. Specifically, the BPD allows a fast image partition into super-BPDs, a powerful alternative of classical superpixels with robust direction similarity. We then construct a region adjacency graph and merge super-BPDs based on the direction similarity along their boundaries, resulting in a fast image segmentation. 
%we adopt a simple CNN to learn the BPDF, based on which we first partition the image into superpixels such that each pixel and its neighboring pixel it points to are in the same superpixel. We then construct the graph of superpixels and define attractive force on each edge as the direction similarity along the boundaries of neighboring superpixels. Besides, we regard neighboring superpixels having small direction similarities as repulsive edges. A final segmentation is obtained by a simple merging algorithm based on attractive forces and repulsive edges. 
The proposed super-BPD achieves a good compromise between accuracy and efficiency, and can separate nearby regions with weak boundaries. In particular, super-BPD achieves comparable or superior performance with MCG but being near real-time. Besides, super-BPD also has an appealing transferability to unseen scenes. This allows potential use of super-BPD in many vision tasks. For example, we have verified the usefulness of super-BPD in object proposal generation. In the future, we would like to explore super-BPD in other applications. 

%\subsection{Weakness}
%\label{subsec:weakness}
Though the proposed super-BPD achieves a pleasant trade-off between generic image segmentation accuracy and efficiency, it is still difficult for super-BPD to accurately segment small regions. This is because that the prediction of BPD around small regions is not very accurate due to dramatic changes of direction. Some failure cases are illustrated in Fig.~\ref{fig:failurecase}, where small regions are not accurately segmented. The segmentation of very small regions also remains a problem for other image segmentation methods. 

\section*{Acknowledgement}

This work was supported in part by the Major Project for New
Generation of AI under Grant no. 2018AAA0100400,
NSFC 61703171, and NSF of Hubei Province of China
under Grant 2018CFB199.
Dr. Yongchao Xu was supported by
the Young Elite Scientists Sponsorship Program by CAST
% Dr. Xiang Bai by the Program for HUST Academic Frontier Youth Team 2017QYTD08, 
and Dr. Donglai Wei by NSF IIS-1835231.
% This work was supported by NSFC 61703171, and in part by the NSF of Hubei Province of China under Grant 2018CFB199, to Dr. Yongchao Xu by the Young Elite Scientists Sponsorship Program by CAST.

%\newpage
%\newpage
{\small
\bibliographystyle{ieee_fullname}
\bibliography{egbib}
}

\end{document}